\documentclass[10pt,twocolumn,letterpaper]{article}

\usepackage[pagenumbers]{cvpr} 

\definecolor{cvprblue}{rgb}{0.21,0.49,0.74}
\usepackage[pagebackref,breaklinks,colorlinks,allcolors=cvprblue]{hyperref}
\usepackage{makecell}
\usepackage{multirow}
\usepackage{multicol}
\usepackage{graphicx}

\usepackage{marvosym}

\def\paperID{*****} 
\def\confName{CVPR}
\def\confYear{2025}

\title{CamCloneMaster: Enabling Reference-based Camera Control for Video Generation}
\author{
    \textbf{Yawen Luo}$^{1\bigtriangledown}$ \ 
    \textbf{Jianhong Bai}$^{2\bigtriangledown}$ \ 
    \textbf{Xiaoyu Shi}$^3$\textsuperscript{\Letter} \ 
    \textbf{Menghan Xia}$^3$ \ 
    \textbf{Xintao Wang}$^3$ \ 
    \textbf{Pengfei Wan}$^3$ \ \\
    \textbf{Di Zhang}$^3$ \ 
    \textbf{Kun Gai}$^3$ \ 
    \textbf{Tianfan Xue}$^1$\textsuperscript{\Letter} \\
    $^1$The Chinese University of Hong Kong \quad 
    $^2$Zhejiang University \\
    $^3$Kuaishou Technology\quad 
    $\textsuperscript{\Letter}$Corresponding author\\
    \textit{\{yawenluo@cuhk.edu.hk,\enspace xiaoyushi@link.cuhk.edu.hk, \enspace tfxue@ie.cuhk.edu.hk\}}
    \\
    \textcolor{magenta}{\url{https://camclonemaster.github.io/}}
}
\begin{document}

\twocolumn[{%
\renewcommand\twocolumn[1][]{#1}%
\maketitle
\vspace{-1.3cm}
\begin{center}
    \centering
    \captionsetup{type=figure}
    \includegraphics[width=\linewidth]{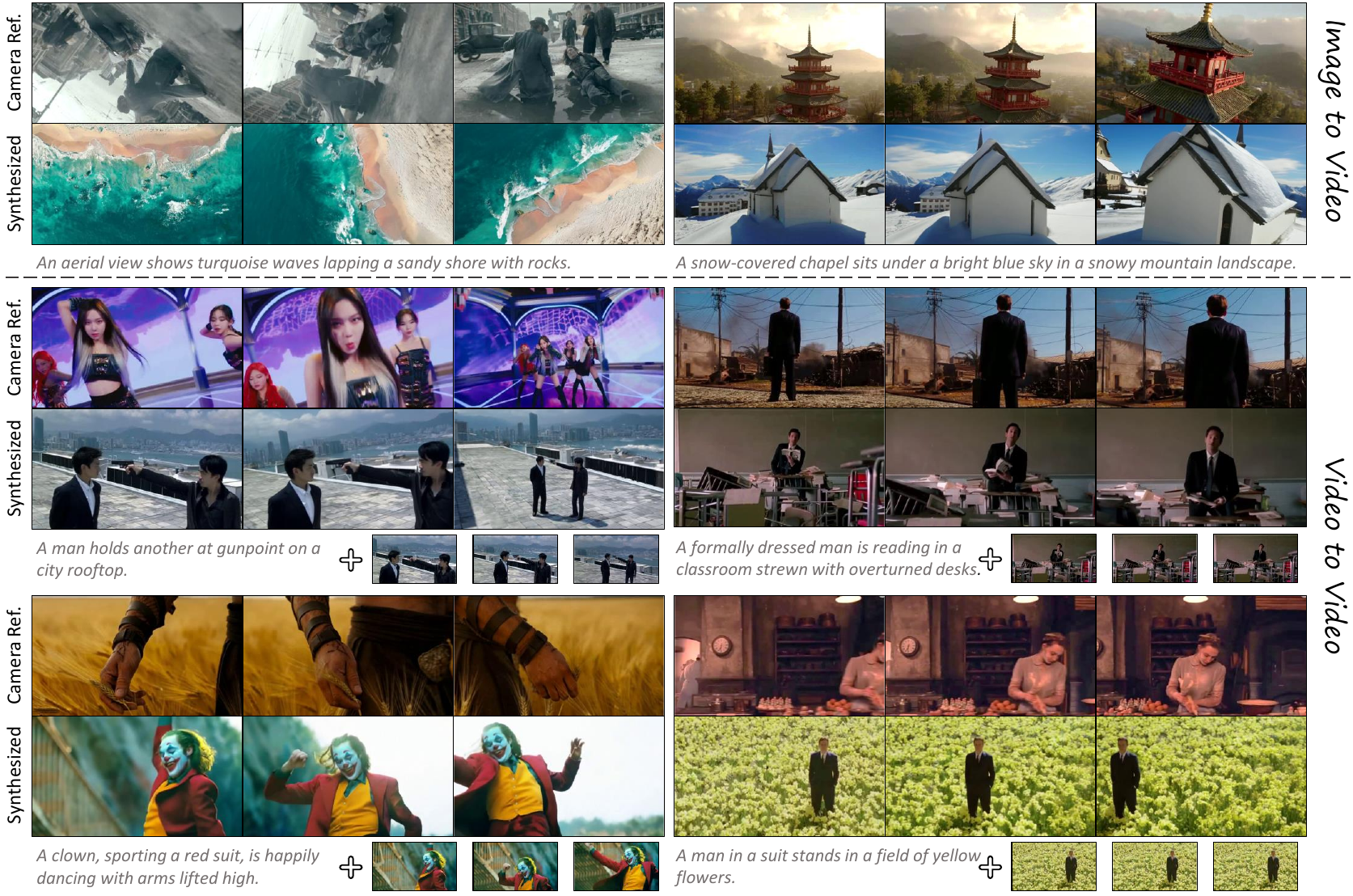}
    \vspace{-15pt}
    \captionof{figure}{Camera Control results of CamCloneMaster. CamCloneMaster is capable of cloning camera motion from reference videos without requiring camera parameters or test-time fine-tuning, which also unifies camera-controlled image-to-video generation and video-to-video re-generation within a single model. For V2V re-generation, the downsized content reference video is positioned beside the prompt. We highly encourage readers to check our demo video for video results, which cannot be well demonstrated by still images.}
    \label{fig:teaser}
\end{center}%
}]

\newcommand\blfootnote[1]{%
\begingroup
\renewcommand\thefootnote{}\footnote{#1}%
\addtocounter{footnote}{-1}%
\endgroup
}

\blfootnote{$^\bigtriangledown$Work done during internship at KwaiVGI, Kuaishou Technology.}

\begin{abstract}
Camera control is crucial for generating expressive and cinematic videos. Existing methods rely on explicit sequences of camera parameters as control conditions, which can be cumbersome for users to construct, particularly for intricate camera movements. To provide a more intuitive camera control method, we propose CamCloneMaster, a framework that enables users to replicate camera movements from reference videos without requiring camera parameters or test-time fine-tuning. CamCloneMaster seamlessly supports reference-based camera control for both Image-to-Video and Video-to-Video tasks within a unified framework. Furthermore, we present the Camera Clone Dataset, a large-scale synthetic dataset designed for camera clone learning, encompassing diverse scenes, subjects, and camera movements. Extensive experiments and user studies demonstrate that CamCloneMaster outperforms existing methods in terms of both camera controllability and visual quality.
\end{abstract}    
\section{Introduction}
\label{sec:intro}

Camera movement is pivotal in video production for creating compelling and dynamic results. It not only helps content creators and cinematographers to frames scene content and dictate global motion, but also to craft specific atmospheres and emphasizes character emotions.
For instance, a push-in shot (moving towards a subject) directs audience attention to details and highlights key moments, whereas a pull-out shot can de-emphasize subjects, detach the audience, and convey negative emotions like isolation. Driven by the growing demand for generating videos with cinematic and expert camera movements, camera controllable video generation~\cite{wang2024motionctrl,xu2024camco,he2025cameractrl,zheng2024cami2v,hu2024motionmaster,ling2024motionclone,hou2025trainingfreecameracontrolvideo} has gained increasing attention.

Existing camera control in video generation often requires explicit camera parameters~\cite{hu2024motionmaster,he2025cameractrl,zheng2024cami2v,li2025realcami2v,bahmani2025ac3d,bahmani2025vd3d,bai2025recammaster,yu2025trajectorycrafter}, which are both hard for users to create and suffer from pose inaccuracy. Since videos inherently embed camera movement information, an ideal way for users to control camera motion is to provide a camera motion reference video. For example, to replicate an iconic tracking shot from Titanic, users just provide the original clip as a reference, and the video generator shall mimic that camera motion.

However, achieving this reference-based camera control is not easy. A naive approach is to first estimate camera parameter sequences from the reference video and then to inject these inference results into camera parameter-based generation models. However, this direct solution faces two major problems: 1) Accurate camera parameters estimation is difficult. Precisely extracting camera parameters from dynamic videos is a challenging problem~\cite{li2024megasam,zhang2024monst3r,wang2024dust3r}. Errors in these inferred parameters inevitably degrade the controllability of parameter-based methods, \textit{i.e.}, the fidelity of camera control is capped by the performance of the camera pose estimation model~\cite{li2024megasam,wang2024dust3r,zhang2024monst3r}. 2) The pose estimation may introduce high interaction cost and computational overhead. A previous training-free method, MotionClone~\cite{ling2024motionclone}, achieves a reference-based control by utilizing sparse temporal attention weights as motion representations for guidance. However, an inversion process to assess these weights introduces additional inference overhead and is vulnerable to unreliable guiding priors.

To address these issues, we propose CamCloneMaster, a novel training-based framework that directly clones camera motion from reference videos. It does not require explicit camera parameters, nor the costly test-time fine-tuning. The user can easily specify the desirable camera motion through a reference video, which eliminates the camera pose estimation stage.

To learn camera motion from reference videos as guidance, our model employs a simple yet effective design: directly concatenating condition tokens with noisy video tokens in a unified input sequence. This design is both parameter-efficient and avoids extra control modules. Moreover, this architecture allows CamCloneMaster to seamlessly support both camera-controlled image-to-video (I2V) generation and video-to-video (V2V) re-generation within a single model. For V2V tasks, users specify an additional content reference, and CamCloneMaster will re-shoot this video using the reference camera motion, as shown in the bottom row of Fig.~\ref{fig:teaser}, providing a powerful post-capture editing tool for cinematographers.

Due to the lack of datasets consisting of paired videos with identical camera trajectories or dynamic scenes for camera clone learning, we construct the Camera Clone Dataset using Unreal Engine $5$~\cite{UnrealEngine5}. This is a large-scale, high-quality dataset, consisting of $391$K realistic videos from $39.1$K distinct locations across $40$ diverse scenes, incorporating $97.75$K diverse camera trajectories. These $40$ scenes cover various environments (indoors/outdoors, day/night) and include carefully curated characters with various motions to simulate real-world complexity. Furthermore, we establish a comprehensive set of rules to automatically generate these realistic and varied camera trajectories, ranging from basic to complex movements.

At last, we evaluate CamCloneMaster through quantitative and qualitative experiments on the RealEstate10K test set~\cite{zhou2018realestate10k} and a curated collection of classic movie clips exhibiting complex camera trajectories. The results demonstrate the advantages of our proposed CamCloneMaster, achieving state-of-the-art performance in camera accuracy, visual quality, and dynamic quality. To further assess the quality of generation results from a subjective aspect, we also conduct a user study involving $47$ participants on $24$ camera motion reference videos collected from the internet. The comparison with baselines reveals a preference among users for the camera control accuracy and visual quality of videos generated by our model.

In summary, our main  contributions are as follows:

\begin{itemize}
\item We introduce CamCloneMaster, a novel framework enabling precise, reference-based camera control for video generation. It operates without camera parameters or test-time fine-tuning, offering a convenient and intuitive user experience.
\item CamCloneMaster uniquely integrates camera-controlled I2V generation and V2V re-generation within a single model using token concatenation—a simple and efficient method that eliminates the need for additional control modules.
\item We construct the Camera Clone Dataset for camera clone learning: a large-scale, high-quality collection of paired videos with identical camera trajectories and dynamic scenes. This dataset will be publicly released to advance future research.
\end{itemize}

\section{Related Work}
\label{sec:related_work}

\noindent\textbf{Video Generation.} The field of video generation is rapidly advancing, with numerous models~\cite{Sora,hong2022cogvideo,ho2022videodiffusionmodels,wanteam2025wan2.1,he2023latentvideodiffusionmodels,ma2025latte} developed to synthesize videos from text. Inspired by the high-fidelity outputs of text-to-image models like Stable Diffusion~\cite{rombach2022stablediffusion} and Flux~\cite{flux}, a parallel research stream explores Image-to-Video (I2V) synthesis~\cite{hong2022cogvideo,wanteam2025wan2.1,xing2023dynamicrafter,zhang2023i2vgenxl,shi2024motioni2v}. For instance, CogVideo~\cite{hong2022cogvideo} and SVD~\cite{blattmann2023stablevideodiffusion} condition generation by concatenating the first frame's latent representation with noise along the channel dimension. Wan-I2V~\cite{wanteam2025wan2.1} extends this by padding and VAE-compressing the conditional frame, adding a positional binary mask, and then concatenating these with noise channel-wise.

\noindent\textbf{Dynamic Videos Camera Parameters Estimation.} Several methods address camera parameter estimation in dynamic videos. A common strategy, employed by Particle-SfM~\cite{zhao2022particlesfm}, LEAP-VO~\cite{chen2024leapvo}, and MegaSam~\cite{li2024megasam}, is to distinguish dynamic from static zones, thereby down-weighting the contributions of dynamic features during inference. Alternatively, MonST3R~\cite{zhang2024monst3r} leverages a 3D point cloud representation, localizing the camera via an additional alignment optimization. Despite their effectiveness, these methods can falter in complex scenarios, such as intricate camera trajectories or scenes dominated by dynamic objects.

\noindent\textbf{Camera Controllable Video Generation.} Camera-controllable video generation methods could be categorized by their need for explicit camera parameters. The first category requires such parameters~\cite{wang2024motionctrl,xu2024camco,he2025cameractrl,li2025realcami2v,zheng2024cami2v,he2025cameractrlii,hou2025trainingfreecameracontrolvideo,song2025lightmotion,wang2025cinemaster}. For instance, MotionCtrl~\cite{wang2024motionctrl} injects extrinsic matrices in temporal attention layers for perspective control, while CameraCtrl~\cite{he2025cameractrl} employs Plücker embedding~\cite{sitzmann2022pluckerembedding} for richer geometric information.  CamCo~\cite{xu2024camco} and CamI2V~\cite{zheng2024cami2v} utilize epipolar attention to enforce geometric constraints. While these methods achieve effective camera control, a key limitation is that obtaining and specifying explicit camera parameters can be difficult and inconvenient for users. The other category~\cite{hu2024motionmaster,wang2024motionctrl} is parameter-free. MotionMaster~\cite{hu2024motionmaster} and MotionClone~\cite{ling2024motionclone} employ an inversion process to derive motion representations from temporal attention maps. However, these methods often exhibit limited generalization and can struggle in complex scenarios.

\noindent\textbf{Camera Controllable Video-to-Video Re-Generation.} Camera controllable V2V re-generation~\cite{vanhoorick2024gcd,gu2025das,bai2025recammaster,yu2025trajectorycrafter,zhang2024recapture,bian2025gsdit} re-shoots the dynamic scenes from a content reference using specified camera trajectories. GCD~\cite{vanhoorick2024gcd} pioneers this task with Kubric-simulated data, limiting its generalization to real-world scenes. DaS~\cite{gu2025das}, GS-Dit~\cite{bian2025gsdit} and Trajectory-Attention~\cite{yu2025trajectorycrafter} utilize $3$D point tracking to extract dynamic information, which can cause artifacts if tracking fails. Though effective, these methods all require accurate camera parameters, which are often difficult and inconvenient for users to provide.

\section{CamCloneMaster}
\label{sec:camclonemaster}

\begin{figure*}[!t]
  \centering
  \includegraphics[width=\linewidth]{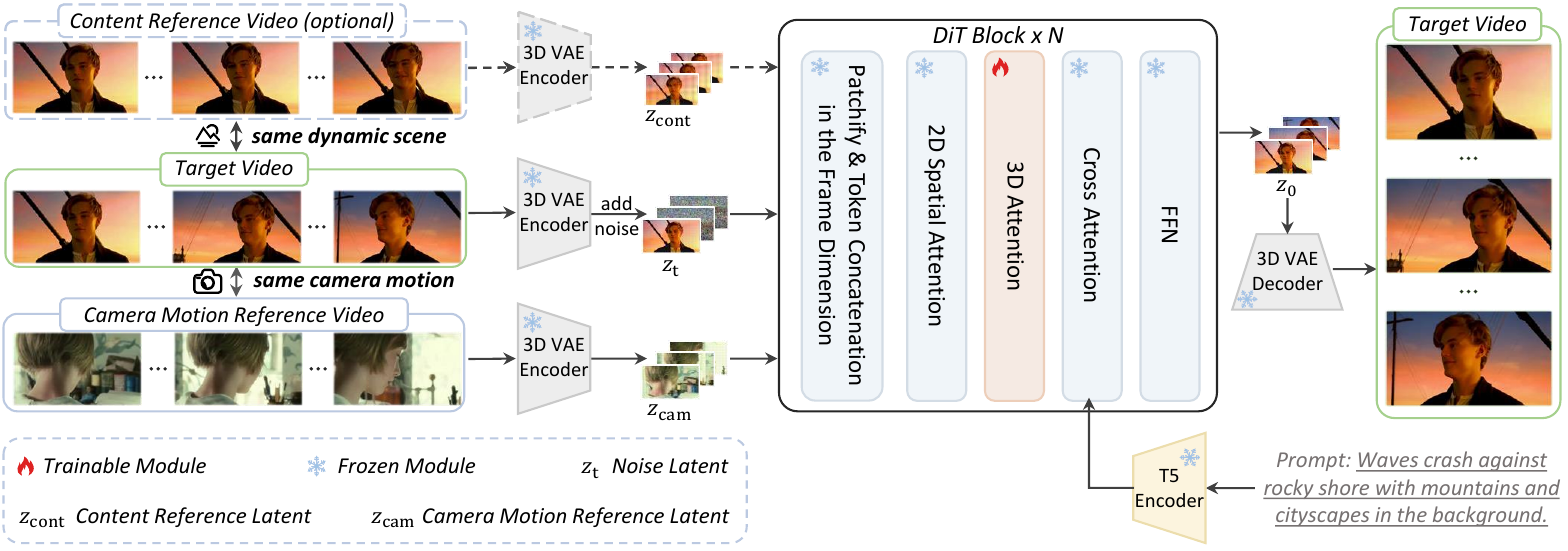}
  \caption{Overview of our proposed CamCloneMaster. Given a camera motion reference video and an optional content reference video as inputs, $3$D VAE encoder is utilized to convert reference videos into conditional latents $z_\textrm{{cam}}$ and $z_\textrm{{cont}}$. We inject the conditional latents into the model by concatenating them with the noise latent along the frame dimension. And only $3$D spatial-temporal attention layers in DiT Blocks are trainable modules in the training process.}  
  \label{fig:network_framework}
  \vspace{-10pt}
\end{figure*}

In this section, we detail the design of our proposed CamCloneMaster. We first describe the components of the base model (Sec.~\ref{sec:preliminary}). Next, we explain our method for extracting camera motion from reference videos as guidance (Sec.~\ref{sec:reference_video_inject_mechanism}). Finally, we introduce CamCloneMaster's training strategy (Sec.~\ref{sec:training_strategy}).

\subsection{Preliminary: Base Model}
\label{sec:preliminary}
Our proposed model, CamCloneMaster, builds upon a transformer-based latent diffusion architecture. This architecture comprises a $3$D Variational Auto-Encoder (VAE)~\cite{kingma2022vae} for latent space mapping and a series of transformer blocks for sequence modeling. Each basic transformer block consists of $2$D spatial self-attention, $3$D spatial-temporal attention, cross-attention, and feed-forward network (FFN). The text prompt embedding $c_\textrm{text}$ is obtained by T$5$ encoder $\varepsilon_{T5}$~\cite{raffel2023t5embedding} and injected into model through cross-attention. We adopt the Rectified Flow~\cite{liu2022flowstraightfastlearning,lipman2023flowmatchinggenerativemodeling} framework to train the diffusion transformer, such that we can generate data sample $\boldsymbol{x}$ from a starting Gaussian sample $\boldsymbol{z}\in \mathcal{N}(\boldsymbol{0},\boldsymbol{I})$. 
Specifically, for a data point $\boldsymbol{x}$, we construct its noised version $\boldsymbol{x}_t$ at timestep $t$ as
\begin{equation}
\boldsymbol{x}_t = (1 - t)\boldsymbol{x} + t\boldsymbol{z}.
\end{equation}
The training objective is a simple MSE loss:
\begin{equation}
\mathcal{L}_{RF}(\theta)= \mathbb{E}_{t,\boldsymbol{x},\boldsymbol{z}}\left \| \boldsymbol{v}_{\theta}(\boldsymbol{x}_t, t, \boldsymbol{c}_I, \boldsymbol{c}_\textrm{text}) - (\boldsymbol{z}-\boldsymbol{x}) \right \|_2^2,
\end{equation}
where the velocity $\boldsymbol{v}_\theta$ is parameterized by the network $\theta$.

\subsection{Reference Videos Injection via Token Concatenation}
\label{sec:reference_video_inject_mechanism}
CamCloneMaster is designed to replicate camera movement from a camera motion reference video $V_\textrm{{cam}}$. Our model directly conditioning on $V_\textrm{{cam}}$, obviates the need for separate camera pose estimation, which not only improves user convenience but also mitigates the risk of pose estimation failures. For V2V re-generation, CamCloneMaster can further incorporate a content reference video $V_\textrm{{cont}}$, enabling it to re-shoot dynamic scenes from $V_\textrm{{cont}}$ while precisely adhering to the $V_\textrm{{cam}}$'s camera movements, as shown in Fig.~\ref{fig:teaser}.

To inject reference camera motion video $V_\textrm{{cam}}$, CamCloneMaster employs a straightforward and efficient design through token concatenation. Specifically, as shown in Fig.~\ref{fig:network_framework}, it merges condition tokens with noisy video tokens by concatenating them along the frame dimension into a single input sequence. This approach is parameter-efficient and eliminates the need for additional modules. 

Another advantage of token concatenation is its ability to support both camera-controlled image-to-video generation and video-to-video re-generation within a single, unified framework. For encoding these reference inputs, CamCloneMaster reuses the 3D VAE $\varepsilon$ from the base model to transform $V_\textrm{{cam}}$ (and $V_\textrm{{cont}}$, in V2V re-generation) into conditioning latents,
\begin{equation}
    z_{i} = \varepsilon(V_{i}),V_i\in \{V_\textrm{{cam}},V_\textrm{{cont}}\}\,, 
\end{equation}
where the $z_{i} \in \mathbb{R}^{f\times c\times h \times w}$, with $f$ frames, $c$ channels, and $h \times w$ spatial size (Fig.~\ref{fig:network_framework}). 

Building upon the shared latent space, we integrate condition latents through token concatenation. As shown in Fig.~\ref{fig:network_framework}, we first patchify the condition latents and the noisy latent $z_{t}$ into tokens:
\begin{equation}
    x_{j} = \text{Patchify}(z_{j}),z_j\in \{z_\textrm{{cam}},z_{\textrm{cont}},z_t\}\,, 
\end{equation}
In I2V setting, $z_\textrm{{cont}}$ will be replaced with all-zero latent. Then, condition tokens $x_\textrm{{cam}}$, $x_\textrm{{cont}}$, and video token $x_t$ are concatenated along the frame dimension:
\begin{equation}
    x_\textrm{{input}} = \textrm{Frame\_Concat}(x_t, x_\textrm{{cam}}, x_\textrm{{cont}})\,,
\end{equation}
where $x_\textrm{{input}}$ is the input of diffusion transformer blocks. The notation $\textrm{Frame$\_$Concat}$ denotes that the condition tokens are concatenated with the noise token in the frame dimension. This design enables the DiT's 3D spatial-temporal attention layers to directly model interactions between condition and noise tokens, without introducing new layers or parameters to the base model.

\subsection{Training Strategy}
\label{sec:training_strategy} Our goal is to finetune the model for camera motion cloning from reference videos while retaining its fundamental generative capabilities. For efficiency and preservation of these capabilities, we selectively finetune only the 3D spatio-temporal attention layers within the DiT blocks. To equip a single model with both image-to-video and video-to-video capabilities, we implement a balanced training approach with $50\%$ camera-controlled image-to-video generation and $50\%$ video-to-video re-generation.

\section{Camera Clone Dataset}
\label{sec:camera_clone_dataset}

\begin{figure*}[!t]
  \centering
  \includegraphics[width=\linewidth]{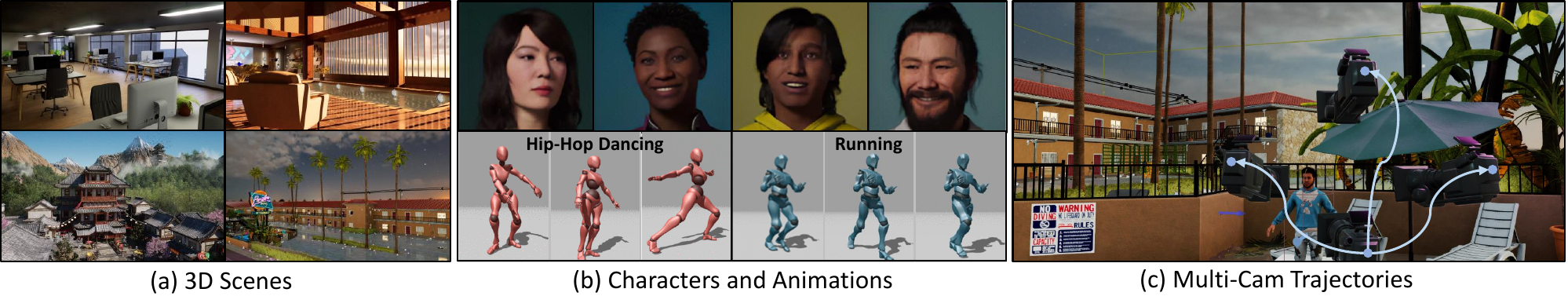}
  \caption{Dataset Construction Illustration. We collect several $3$D scenes as background, and put characters into scenes as foreground, each character is combined with a specific animation. Then, multiple paired camera trajectories are designed and shots are made by rendering in Unreal Engine 5.}
  \label{fig:ue_data_pipeline}
\end{figure*}

\begin{table*}[!t]
\begin{center}
\caption{Quantitative Results for Camera-Controlled Image-to-Video Generation and Video-to-Video Generation. The best performance is in \textbf{boldface}, while the second is \underline{underlined}. Sub. Cons. and Bg. Cons. denote Subject Consistency and Background
Consistency, respectively, as defined in Sec.~\ref{sec:experiment_setup}, here and after.}
\label{table:Quantitative_i2v}
\vspace{-4pt}
\setlength\tabcolsep{2pt}
\resizebox{0.995\textwidth}{!}{
    \begin{tabular}{lc|cccc|ccc|cccc}
        \toprule
        \multirow{2}{*}{Method} & \multirow{3}{*}{\makecell[c]{Camera\\Input}} & \multicolumn{4}{c|}{Visual Quality} & \multicolumn{3}{c|}{Camera Accuracy} & \multicolumn{4}{c}{Dynamic Quality}\\ 
            \cmidrule(r){3-13}
            & &\makecell[c]{ \makecell[c]{Imaging \\ Quality} $\uparrow$} & \makecell[c]{\makecell[c]{CLIP \\ Score} $\uparrow$}  & FVD$\downarrow$ & FID$\downarrow$ & \makecell[c]{Rot\\Err} $\downarrow$& \makecell[c]{Trans\\Err} $\downarrow$ & \makecell[c]{Cam\\MC} $\downarrow$ &\makecell[c]{\makecell[c]{Dynamic \\ Degree} $\uparrow$} & \makecell[c]{\makecell[c]{Motion \\ Smooth} $\uparrow$} & \makecell[c]{\makecell[c]{Sub. \\ Cons.} $\uparrow$}& \makecell[c]{\makecell[c]{Bg. \\ Cons.}
            $\uparrow$}\\
            \midrule 
            \multicolumn{13}{c}{Camera-Controlled Image-to-Video Generation} \\
            \midrule 
             CameraCtrl~\cite{he2025cameractrl}& Cam. Param.&60.47& 19.81 & 1294.97 & 173.85 & 2.82 & 4.52 & 6.48 &30.62 & 89.39 & 80.34 & 89.29 \\
             CamI2V~\cite{zheng2024cami2v}& Cam. Param.&62.88 & 22.13 &  \underline{1013.23} & \underline{101.52} & \underline{1.62} & \underline{3.07} & \underline{4.22} & 27.15&\underline{89.90} & \underline{87.15} & \underline{89.99} \\
             MotionClone~\cite{ling2024motionclone}&Ref. Video& \underline{64.14} &\underline{25.04} & 1355.55 &191.43 & 3.10 & 5.15& 7.31 & \underline{47.03} &  81.42& 70.99 & 79.06 \\
            \midrule
            CamCloneMaster&Ref. Video& \textbf{64.65}&  \textbf{25.08}&  \textbf{993.06}&\textbf{99.96} & \textbf{1.49} & \textbf{2.37}& \textbf{3.50} &\textbf{50.11} &  \textbf{94.29}  & \textbf{92.78} & \textbf{93.86} \\
            \midrule             
            \multicolumn{13}{c}{Camera-Controlled Video-to-Video Re-Generation} \\
            \midrule 
             DaS~\cite{gu2025das}& Cam. Param.& \underline{62.07}& \underline{22.91} &721.68 & \underline{69.95} & 2.71 & 8.18 & 9.62 &  23.54&93.91& \underline{93.38} & \underline{93.77}\\
             ReCamMaster~\cite{bai2025recammaster}&Cam. Param.&53.71 & 22.48 & \underline{718.69} & 101.82& \underline{1.53} & \underline{3.12} & \underline{4.17} &  \underline{58.57}&  88.13&  87.59 & 88.82\\
             TrajectoryCrafter~\cite{yu2025trajectorycrafter}&Cam. Param.& 61.92& 21.58& 1086.89  & 132.47&  3.08& 7.46 & 10.22 & 54.92 & \underline{96.18} & 90.07& 86.09 \\
            \midrule
            CamCloneMaster&Ref. Video &\textbf{ 62.78} & \textbf{24.15} & \textbf{678.06} & \textbf{60.03}& \textbf{1.36} &\textbf{2.02} &\textbf{3.05} & \textbf{60.19} & \textbf{98.97} & \textbf{94.71} & \textbf{94.32}  \\
        \bottomrule
    \end{tabular}}
    \vspace{-20pt}
\end{center}
\end{table*}

Reference-based camera clone learning requires triple video sets: a camera motion reference video $V_\textrm{{cam}}$, a content reference video $V_{\textrm{cont}}$, and a target video $V$, which recaptures the scene in $V_{\textrm{cont}}$ with the same camera movement as $V_{\textrm{cam}}$. Building such a dataset in the real world is difficult and label-intensive. Therefore, we choose to build our camera clone dataset by rendering it in Unreal Engine 5~\cite{UnrealEngine5}. As illustrated in Fig.~\ref{fig:ue_data_pipeline}, we collect $40$ 3D scenes as backgrounds. And we also collect $66$ characters and put them into the $3$D scenes as main subjects, each character is combined with one random animation, such as running and dancing.

To construct the triple set, camera trajectories must satisfy two key requirements: 1) \textit{Simultaneous Multi-View Capture}: Multiple cameras must film the same scene concurrently, each following a distinct trajectory. 2) \textit{Paired Trajectories}: paired shots with the same camera trajectories across different locations. Our implementation strategy addresses these needs as follows: Within any single location, 10 synchronized cameras operate simultaneously, each following one of ten unique, pre-defined trajectories to capture diverse views. To create paired trajectories, we group $3$D locations in scenes into sets of four, ensuring that the same ten camera trajectories are replicated across all locations within each set. The camera trajectories themselves are automatically generated using designed rules. These rules encompass various types, including basic movements, circular arcs, and more complex camera paths.

In total, our camera clone dataset comprises $391$K visually authentic videos shooting from $39.1$K different locations in $40$ scenes with $97.75$K diverse camera trajectories, and $1,155$K triple video sets are constructed based on these videos. Each video has a resolution of $576\times1,008$ and $154$ frames.
\section{Experiments}
\subsection{Experiment Setup}
\label{sec:experiment_setup}
\begin{figure*}[!t]
  \centering
  \includegraphics[width=\linewidth]{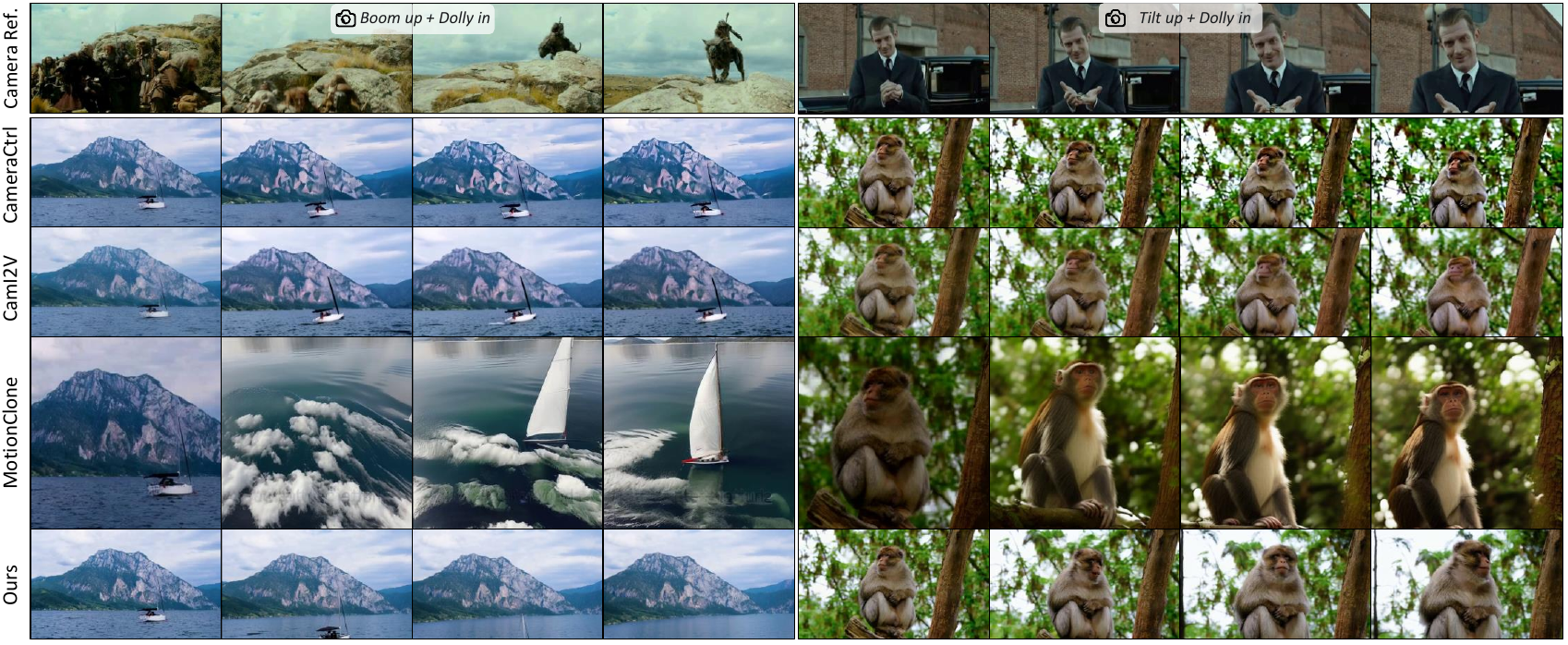}
  \vspace{-10pt}
  \caption{Quantitative Results for Camera-Controlled Image-to-Video Generation. Camera poses are estimated using MegaSam for parameter-based methods.}
  \label{fig:exp_i2v}
    \vspace{-10pt}
\end{figure*}
\textbf{Implement Details.} CamCloneMaster is trained based on an internal image-to-video diffusion model using the rendered Camera Clone Dataset (detailed in Sec.~\ref{sec:camera_clone_dataset}). For training, each video is resized to a resolution of $384\times672$ and uniformly sampled into $77$ frames. We optimize only the 3D spatial-temporal attention layers within the DiT blocks using the Adam optimizer with a learning rate of $5e-5$. The model is trained for $12,000$ steps on a cluster of $64$ NVIDIA H800 GPUs, utilizing a batch size of $64$. 

\noindent\textbf{Evaluation Set.} For camera motion references, we randomly selected 1,000 videos from the RealEstate10K~\cite{zhou2018realestate10k} test set. These videos offer $1,000$ camera trajectories and are annotated with camera parameters, which are leveraged as conditional inputs for camera parameter-dependent methods. For content references, another $1,000$ videos were randomly chosen from Koala-36M~\cite{wang2025koala36m}. In the camera-controlled image-to-video setting, only the first frame of each content video is used as a conditional input.

\noindent\textbf{Evaluation Metrics.} (1) \textit{Visual Quality}: The quality of the synthesized videos is evaluated using Imaging Quality~\cite{huang2023vbench}, Clip Score~\cite{huang2023vbench}, Fr$\acute{e}$chet Video Distance (FVD)~\cite{unterthiner2019fvd}, and Fr$\acute{e}$chet Inception Distance (FID)~\cite{heusel2018fid}. (2) \textit{Dynamic Quality}: To evaluate video dynamics and temporal consistency, we adapt VBench~\cite{huang2023vbench} metrics. Dynamic Degrees and Motion Smoothness assess motion range and temporal coherence, while Subject and Background Consistency evaluate foreground and background temporal consistency, respectively. (3) \textit{Camera Accuracy}: To evaluate camera trajectory accuracy, we use the state-of-the-art camera parameters estimation model MegaSaM~\cite{li2024megasam} to extract camera rotation $R_i$ and translation $T_i$ from the $i$ frame in the video. We then calculate the Rotation Distance (RotErr), Translation Error (TransErr), and Camera Motion Consistency (CamMC), following CamI2V~\cite{zheng2024cami2v}. (4) \textit{View Consistency}: To evaluate dynamic scene consistency in camera-controlled V2V generation against content references, we following ReCamMaster~\cite{bai2025recammaster}. We use GIM~\cite{shen2024gim} for Matching Pixels (pixels matched above a confidence threshold), and compute FVD-V~\cite{xie2025sv4d} (FVD between the reference and generated) and CLIP-V~\cite{bai2025recammaster} (frame-wise CLIP similarity). 


\begin{figure*}[!t]
  \centering
  \includegraphics[width=\linewidth]{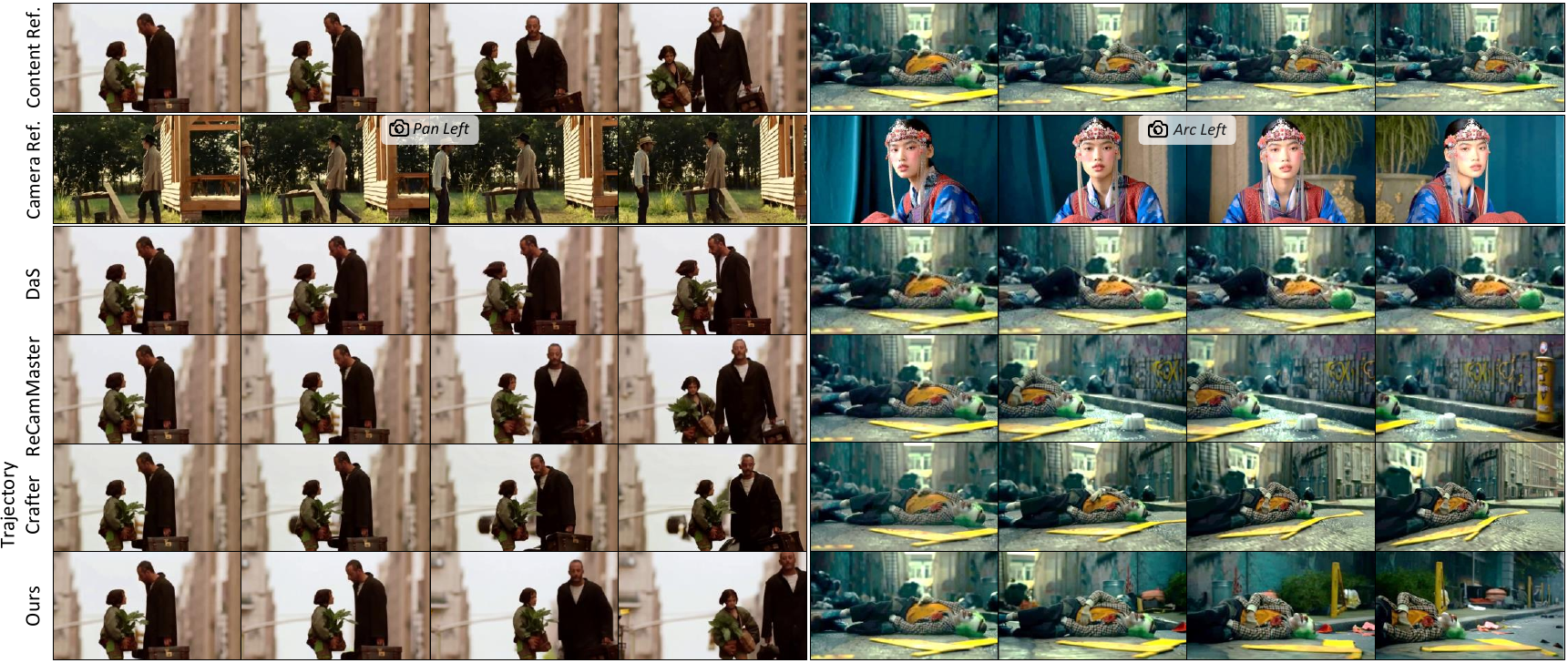}
  \vspace{-10pt}
  \caption{Quantitative Results for Camera-Controlled V2V Re-Generation. Camera poses are estimated using MegaSam for parameter-based methods.}
  \label{fig:v2v}
  \vspace{-10pt}
\end{figure*}

\subsection{Comparisons with State-of-the-Art Methods}
\subsubsection{Camera Controlled Image-to-Video Generation}
\textbf{Baselines.} We compare our proposed CamCloneMaster with state-of-the-art camera-controlled image-to-video generation methods~\cite{he2025cameractrl,zheng2024cami2v,ling2024motionclone}. CameraCtrl~\cite{he2025cameractrl} and CamI2V~\cite{zheng2024cami2v} adopt Plücker embedding as camera representation, while MotionClone~\cite{ling2024motionclone} is a training-free framework enabling cloning motion from reference video directly, which utilizes sparse temporal attention weights as motion representations for motion guidance in the generated process. Although MotionClone does not need camera parameters as inputs, it struggles to handle complex camera movements effectively. 

\noindent\textbf{Quantitative Results.} We validate CamCloneMaster on the test set introduced in Sec.~\ref{sec:experiment_setup}. Our method and MotionClone both directly utilizes the camera motion reference video as condition. For CameraCtrl and CamI2V, which require explicit camera parameters as condition inputs, we provide the ground truth camera poses. Quantitative results in Table~\ref{table:Quantitative_i2v} highlight CamCloneMaster's superior camera control ability and accuracy over competing methods. This is demonstrated by significantly lower RotErr, TransErr, and the CamMC score. Additionally, CamCloneMaster also achieves better visual and dynamic quality compared to the other approaches.      

\noindent\textbf{Qualitative Results.} The qualitative result is shown in Fig.~\ref{fig:exp_i2v}, our method accurately clones camera motion from the reference and achieves high visual quality. For instance, it successfully maintains the detailed structure of the sailboat (left case) and captures the monkey's complex motion (right case). In contrast, CameraCtrl and CamI2V struggle to follow complex trajectories, such as the leftward trucking combined with rotation in the left example. Moreover, MotionClone's limited generalizability restricts its ability to replicate camera motion from reference videos, which also fails to maintain subject consistency. 

\subsubsection{Camera Controlled Video-to-Video Re-generation}
\textbf{Baselines.} We compare CamCloneMaster with DaS~\cite{gu2025das}, ReCamMaster~\cite{bai2025recammaster}, and TrajectoryCrafter~\cite{yu2025trajectorycrafter}, all of which require camera parameters as input. Das~\cite{gu2025das} utilize $3$D point tracking to extract dynamic information from the content reference video, while ReCamMaster employs a video conditioning mechanism. TrajectoryCrafter constructs a point cloud from the content reference and renders new viewpoints, which are then utilized as control signals in the generation process.

\begin{table}[!t]
	\begin{center}
		\caption{Quantitative Results for Camera-Controlled Video-to-Video Re-generation on View Consistency. The best performance is in \textbf{boldface}, while the second is \underline{underlined}.}
        \label{table:Quantitative_v2v_view_consistency}
		\setlength\tabcolsep{4.0pt}
		\begin{tabular}{lccc}
			\toprule
			\multirow{2}{*}{Method} & \multicolumn{3}{c}{\makecell[c]{View  Consistency}} \\ 
                \cmidrule(r){2-4}
                & \makecell[c]{\makecell[c]{Matching \\ Pixels} $\uparrow$} & \makecell[c]{FVD-V $\downarrow$} & \makecell[c]{CLIP-V $\uparrow$}  \\
                \midrule 
                 DaS~\cite{gu2025das} & 969.37 & \underline{182.34}  & \underline{88.44} \\
                 ReCamMaster~\cite{bai2025recammaster}&  \underline{1268.10} &  250.65 & 86.56  \\
                \midrule
                CamCloneMaster& \textbf{1332.34} & \textbf{176.02} & \textbf{88.77} \\
			\bottomrule
		\end{tabular}
	\end{center}
    \vspace{-14pt}
\end{table}

\noindent\textbf{Quantitative Results.} As shown in Table~\ref{table:Quantitative_i2v} and Table~\ref{table:Quantitative_v2v_view_consistency}, CamCloneMaster outperforms baselines across multiple metrics. Our method not only accurately controls the camera and achieves high visual quality but also effectively preserves the dynamic scene from the content reference.

\noindent\textbf{Qualitative Results.} Qualitative results are presented in Fig.~\ref{fig:v2v}. Baseline methods generally fail to generate videos with accurate camera movements. Specifically, DaS and TrajectoryCrafter produce content containing notable artifacts. In contrast, our method accurately clones camera motion from the reference video and creates outputs with high visual quality and temporal consistency.

\subsection{User Study}
We conduct a user study to highlight the importance of accurate camera poses and the challenge of obtaining them for parameter-based methods~\cite{zheng2024cami2v,he2025cameractrl,gu2025das,bai2025recammaster,yu2025trajectorycrafter}. Participants need to compare paired videos: one generated using ground-truth camera parameters, and the other using parameters inferred by the state-of-the-art camera pose estimation model, MegaSam~\cite{li2024megasam}. Users are asked to select which video's camera movement more closely matched the corresponding reference. Specifically, we randomly select $12$ camera motion reference videos, each labeled with GT camera parameters, from our synthetic Camera Clone Dataset. This experiment is conducted based on three parameter-based methods: CamI2V~\cite{zheng2024cami2v}, CameraCtrl~\cite{he2025cameractrl}, and Recammaster~\cite{bai2025recammaster}. The user study involves $47$ participants, and the results presented in Table~\ref{table:use_study_1} indicate a significant preference for videos generated with GT camera parameters over those using MegaSam-inferred parameters. These results demonstrate that: 1) 
Camera movement fidelity in parameter-based methods is highly dependent on input parameters accuracy. 2) Even SOTA pose estimation models struggle to provide sufficiently accurate parameters, motivating our proposed reference-based camera control framework.

\begin{table}[!t]
	\begin{center}
		\caption{User Study. To demonstrate the importance of accurate camera parameters, we generated videos using parameter-based baselines, employing two sets of parameters: 1) ground truth (GT) and 2) those estimated by MegaSam, a state-of-the-art camera pose estimation model. Users are tasked with selecting the video whose camera movement more closely matches a reference video, and we report the resulting preference rates.}
		\label{table:use_study_1}
            \vspace{-4pt}
		\setlength\tabcolsep{2.0pt}
            \resizebox{0.478\textwidth}{!}{
		\begin{tabular}{lccc}
			\toprule
			 Camera Pose& CamI2V & CameraCtrl& ReCamMaster\\ 
                \midrule 
                \makecell[l]{ Ground Truth}&$\textbf{77.67\%}$& $\textbf{87.67\%}$& $\textbf{76.33\%}$\\
                \makecell[l]{Estimated by MegaSaM}&$22.33\%$& $13.33\%$& $23.66\%$\\
			\bottomrule
		\end{tabular}}
	\end{center}
    \vspace{-16pt}
\end{table}

\begin{table}[!t]
	\begin{center}
		\caption{User Study. To validate the advantages of the proposed CamCloneMaster, we collect $24$ videos with complex camera movements from the internet to serve as camera motion reference. Users are asked to select their preferred video from a randomly ordered set of results generated by our method and baseline approaches. Multiple selections are permitted, and the outcomes are presented as user preference rates. }
        \vspace{-2pt}
            \label{table:user_study_2}
		\setlength\tabcolsep{1.5pt}

		\begin{tabular}{lccc}
			\toprule
			Method & \makecell[c]{Camera \\ Accuracy} & \makecell[c]{Video-Text\\Consistency} & \makecell[c]{Temporal \\ Consistency}  \\
                \midrule 
                \multicolumn{4}{c}{Camera-Controlled Image-to-Video Generation} \\
                \midrule 
                 CameraCtrl~\cite{he2025cameractrl} & $10.00\%$  & $20.00\%$ & $11.88\%$\\
                 CamI2V~\cite{zheng2024cami2v}& $5.00\%$  & $21.88\%$ &$12.50\%$ \\
                 MotionClone~\cite{ling2024motionclone}& $13.13\%$  & $31.88\%$ & $13.75\%$\\
                \midrule
                CamCloneMaster& $\textbf{85.00\%}$ & $\textbf{81.25\%}$ & $\textbf{88.13\%}$ \\
                \midrule
                \multicolumn{4}{c}{Camera-Controlled Video-to-Video Re-Generation} \\
                \midrule 
                 DaS~\cite{gu2025das} & $18.13\%$ & $27.50\%$ & $28.13\%$\\
                 ReCamMaster~\cite{bai2025recammaster}& $16.25\%$ &$27.50\%$  &$34.38\%$ \\
                 TrajectoryCrafter~\cite{yu2025trajectorycrafter}& $8.75\%$ & $4.38\%$ & $4.38\%$\\
                \midrule
                CamCloneMaster& $\textbf{78.75\%}$ & $\textbf{91.88\%}$ &  $\textbf{86.25\%}$\\
			\bottomrule
		\end{tabular}
	\end{center}
       \vspace{-10pt}
\end{table}

Another user study is conducted to better evaluate different methods from the subjective perspective. We collect $24$ camera motion references and $12$ content references from the internet with the resolution of $1080\times1920$. During the test, participants are simultaneously presented with $4$ videos, displayed in a randomized order: one generated by our method and one from each of the three competing baselines relevant to the specific task (I2V and V2V). Participants evaluate the videos on three criteria: 1) Camera Accuracy: how well camera movement followed the camera motion reference video, 2) Video-Text Consistency: how well content aligned with the prompt, and 3) Temporal Consistency. Multiple selections are allowed for each question. The user study involves $47$ participants, and the results in Table~\ref{table:user_study_2} indicate that our method is consistently preferred by most users.

\subsection{Ablation Studies}
\textbf{Ablation on Condition Injection Mechanism.} Our model conditions video generation by concatenating condition tokens with noise latent tokens along the frame dimension. We validate this frame concatenation against the widely used channel concatenation~\cite{yu2025trajectorycrafter,ren2025gen3c}. We also test concatenating conditions solely within the temporal DiT block layers, as explicit attention between condition and noise tokens is limited to the 3D spatio-temporal attention layers. Finally, we compare token concatenation against a ControlNet-like architecture~\cite{wang2025cinemaster}, where duplicated DiT blocks extract reference video features for injection into the base model via feature addition. Result in Table~\ref{tabel:abla_for_injection_method} indicates that concatenating conditions across all layers is essential for optimal performance (Rows 2 and 4).  We reason that global video properties, such as camera motion, demand high-level representations, and even layers lacking explicit attention between the condition and noise token play a vital role in extracting these. Furthermore, token concatenation outperforms the ControlNet-like feature addition (Row 3 and Row 4), likely because feature addition can make it harder for the model to distinguish motion cues from the reference content.
\begin{table}[!t]
	\begin{center}
		\caption{Ablation Study on Condition Injection Mechanism.}
         \vspace{-2pt}
		\label{tabel:abla_for_injection_method}
		\setlength\tabcolsep{2.0pt}
            \resizebox{0.478\textwidth}{!}{
		\begin{tabular}{lccccc}
			\toprule
			 Injection Method& FVD$\downarrow$& \makecell[c]{Rot\\Err}$\downarrow$ & \makecell[c]{Trans\\Err}$\downarrow$& \makecell[c]{Cam\\MC}$\downarrow$& \makecell[c]{\makecell[c]{Bg. \\ Cons.}$\uparrow$}\\ 
                \midrule 
                Channel Concatenation& 1115.83 & 1.68 & 2.88 &3.68&92.30\\
                Only Temporal Layer& 1103.18 &1.76& 3.22 & 3.87&92.06\\
                ControlNet&1808.08& 1.94 & 3.79 & 4.64&93.41\\
                \midrule 
                CamCloneMaster&\textbf{993.06}&\textbf{1.49} & \textbf{2.37}& \textbf{3.50} &\textbf{93.86}\\
			\bottomrule
		\end{tabular}}
	\end{center}
\end{table}

\begin{table}[!t]
	\begin{center}
		\caption{Ablation Study on Training Strategy.}
            \vspace{-2pt}
		\label{tabel:abla_for_training_strategy}
		\setlength\tabcolsep{2.0pt}
            \resizebox{0.478\textwidth}{!}{
		\begin{tabular}{lccccccc}
			\toprule
			 \makecell[c]{Finetune\\ Module} & \makecell[c]{Imaging \\ Quality} $\uparrow$ &\makecell[c]{\makecell[c]{CLIP \\ Score} $\uparrow$}  & \makecell[c]{FVD} $\downarrow$ & \makecell[c]{Rot\\Err}$\downarrow$ & \makecell[c]{Trans\\Err}$\downarrow$ & \makecell[c]{Cam\\MC}$\downarrow$\\ 
                \midrule 
                DiT Block &62.54& 25.06& 1172.46& 1.58 & 2.58 & 3.64 \\
                \midrule 
                $3$D-Attn.&\textbf{64.65}&  \textbf{25.08}&\textbf{993.06}&\textbf{1.49} & \textbf{2.37}& \textbf{3.50}\\
			\bottomrule
		\end{tabular}}
	\end{center}
       \vspace{-10pt}
\end{table}
\noindent\textbf{Ablation on Training Strategy.} We finetune only the 3D spatial-temporal attention layers in DiT blocks and freeze other parameters in the training process. Results in Table~\ref{tabel:abla_for_training_strategy} show that only finetuning 3D spatial-temporal attention layers leads to enhanced camera clone accuracy while also preserving better visual quality.
\section{Conclusion and Limitation}
In this paper, we introduce CamCloneMaster, a novel method for intuitive and user-friendly camera control in video generation. Specifically, CamCloneMaster enables users to replicate camera movements from reference videos without requiring camera parameters or test-time fine-tuning. 
Another innovation is a simple yet effective architecture, which effectively unifies both camera-controlled I2V generation and V2V re-generation within a single model without requiring additional control modules. We also develop a high-quality synthetic dataset for training.

\noindent\textbf{Limitation.} Although the token concatenation strategy shows promising results for camera-controlled video generation, it increases computational demands. Exploring methods like sparse attention and latent drop to mitigate this overhead is reserved for future work.
{
    \small
    \bibliographystyle{ieeenat_fullname}
    \bibliography{main}
}


\end{document}